\documentclass[runningheads]{llncs}
\usepackage[T1]{fontenc}
\usepackage{graphicx}
\usepackage{booktabs}
\usepackage{amsmath}
\usepackage{amssymb}
\usepackage{algorithm}
\usepackage{algpseudocode}
\usepackage{xcolor}

\begin{document}
\title{Meta-Learning for Airflow Simulations with Graph Neural Networks}
%
%
\author{Wenzhuo Liu\inst{1,2}
\and
Mouadh Yagoubi \inst{1}
\and
Marc Schoenauer\inst{2}
}
\authorrunning{W. Liu et al.}
%
\institute{IRT SystemX, Orsay, France 
\email{FirstName.LastName@irt-systemx.fr}
\and
INRIA TAU, LISN CNRS \& U. Paris-Saclay, ORsay, France
}
\maketitle              
\begin{abstract}
The field of numerical simulation is of significant importance for the design and management of real-world systems, with partial differential equations (PDEs) being a commonly used mathematical modeling tool. However, solving PDEs remains still a challenge, as commonly used traditional numerical solvers often require high computational costs. As a result, data-driven methods leveraging machine learning (more particularly Deep Learning) algorithms have been increasingly proposed to learn models that can predict solutions to complex PDEs, such as those arising in computational fluid dynamics (CFD). However, these methods are known to suffer from poor generalization performance on out-of-distribution (OoD) samples, highlighting the need for more efficient approaches. To this end, we present a meta-learning approach to enhance the performance of learned models on OoD samples. Specifically, we set the airflow simulation in CFD over various airfoils as a meta-learning problem, where each set of examples defined on a single airfoil shape is treated as a separate task. Through the use of model-agnostic meta-learning (MAML), we learn a meta-learner capable of adapting to new tasks, i.e., previously unseen airfoil shapes, using only a small amount of task-specific data. We experimentally demonstrate the efficiency of the proposed approach for improving the OoD generalization performance of learned models while maintaining efficiency.

\end{abstract}
%
%
\section{Introduction} 
Nowadays, numerical simulations of partial differential equations (PDEs) are a crucial tool in designing and managing real-world systems. For instance, the Navier-Stokes equations describing the motion of some fluid are commonly used in the analysis of fluid flow systems, such as aerodynamic airfoil design.
However,  solving numerically PDEs remains a challenge despite their widespread use.
Numerical Analysis encompasses different numerical simulation methods that are widely used to approximate the solutions of PDEs. One of the common used methods to deal with this limitation  is the Finite Volume Method (FVM): This approach discretizes the domain into a {\em mesh} and computes approximated values of the quantities of interest in each  cell of the mesh. 
While these approaches can predict the behavior of systems with known error bounds, their computational cost can be extremely high for complex systems, especially in computational fluid dynamics (CFD).

Over the past few years, there has been a significant increase in using machine learning algorithms for numerically solving physical problems, particularly in fluid dynamics, where conventional numerical solvers are computationally expensive.
In such context, data-driven methods have been proposed to learn Deep Network (DN) models that predict the solutions of complex PDEs to make full use of accumulated datasets, either generated by classical solvers, or from experimental sources.
Through Deep Learning (DL) methods, these models can learn from data, allowing for more efficient solutions to PDEs. However, a major limitation of these approaches is their tendency to suffer from poor generalization performance on out-of-distribution (OoD) samples.
As the deep learning methods learn patterns directly from the data instead of studying problem-related constraints beyond the PDEs, they cannot grasp the physics of the problem at hand. The lack of underlying physical laws leads to generalization issues: the predicted results can significantly lack any physical significance, and the learned models often underperform on OoD samples: the deep learning models can predict unseen examples that are close to the training set but perform poorly at solving new problems that significantly differ from the training examples,
as the underlying physical laws are not explicitly incorporated into the learning process.
In order to address this issue, a number of researchers have proposed the use of hybrid models that combine deep learning with computational fluid dynamics (CFD) solvers. \cite{belbuteperes2020combining} takes advantage of the approximated solutions from CFD solvers on coarse meshes, using them as input features for deep learning models to make super-resolution predictions in aerodynamics. \cite{kochkov2021machine} uses machine learning to correct errors in cheap and under-resolved simulations on coarse meshes from traditional solvers.
These models obtain extra information from the coarse-grained simulations to reduce generalization errors.
\cite{obiols2020cfdnet} uses the predictions obtained from deep learning models as initial values and then employs CFD solvers to further refine and improve the solutions from the deep learning models.
While these hybrid models are able to improve the accuracy of predictions on OoD samples, they are often less efficient than pure machine learning methods \cite{belbuteperes2020combining}.  

In this paper, we introduce a meta-learning perspective for CFD tasks and apply an optimization-based meta-learning approach to enhance the performance of the learned models on OoD data points.
Meta-learning provides a way to gain meta-knowledge over various tasks and to use this knowledge to learn new tasks using a few task-specific data points. 
Specifically, we formulate the CFD problem of airflow simulation over various airfoil shapes as a meta-learning problem: learning airflow simulations around one single airfoil is considered as one task. 
By leveraging the information from these tasks with meta-learning, a meta-learner can then adapt to new tasks, i.e., to unseen airfoil shapes, using only a small amount of task-specific sample data points.  

\section{CFD Background}
Computational Fluid Dynamics (CFD) has extensive applications across various fields of study and industries, such as aerodynamics, weather simulation, and heat transfer, among others. The Navier-Stokes equations form the fundamental basis for the majority of CFD problems, governing the behavior of single-phase fluid flows. However, traditional numerical solvers often encounter high computational costs when addressing complex problems in the CFD domain.

In recent years, machine learning algorithms have emerged as a promising alternative for CFD problems, garnering significant interest due to their potential to alleviate computational burdens while maintaining accuracy in simulating fluid dynamics.
\subsection{Machine Learning for CFD}
\subsubsection{Data-driven method} \label{sec: data-driven} 
Several works have proposed data-driven methods, making full use of labeled data accumulated by numerical solvers. The objective is to create a model capable of solving a family of PDEs of the same type. The inputs of the model are the physical quantities that define a specific instance of the target PDE, and the geometrical definition of the domain of interest where this PDE is defined. The 'labels' are the corresponding numerical solutions of the PDE. These methods learn a mapping from the physical and geometrical inputs to the corresponding solutions. Early works in 2D  \cite{Bhatnagar_2019,tompson2017accelerating,thuerey2020deep} considered the problem as an image-to-image problem and applied Convolutional Neural Networks (CNNs) in a straightforward manner. 
However, such a CNN approach only applies to image-like data. These researches either use structured meshes to discretize the physical domain, directly producing image-like data, or apply interpolation to convert mesh data into structured grids. As a result, they could not handle well problems with complex geometry. 

In recent years, many attempts have been made to construct models with Graph Neural Networks (GNNs) to cope with mesh data.  
\cite{kashefi2021point} discusses fluid flow field problems on different irregular geometries, considering CFD data as a set of points (called point clouds) and applying the PointNet \cite{qi2017pointnet} architecture specially designed for such a data type; \cite{pfaff2020learning} proposes a framework for learning mesh-based simulation problems with graph neural networks.
Both these data-driven approaches are extremely efficient compared with numerical solvers in terms of computational cost; however, they always suffer from out-of-distribution generalization issues.  

\subsubsection{Unsupervised learning Approaches} On the other hand, some works \cite{Berg_2018,raissi2018deep,raissi2019physics} based on unsupervised learning aim to design mesh-free methods, e.g., in the case of high dimensions or complex geometry. 
The basic idea of such methods is to put the physics in the loss during training. The inputs of the NN are the coordinates of some points in the domain, and the output is the solution $u$ of the PDE at hand. At inference time, the network directly predicts the value $u(x)$ when given $x$ as input. However, these unsupervised learning methods solve one problem each time (e.g., re-training is necessary when the boundary conditions change) and  require much more computational time for a new problem than data-based approaches.

\subsection{Airfoil Flow Simulation}\label{CFDdimulations}
In this study, we focus on airfoil flow simulation problems in the CFD domain and aim to develop a robust deep learning model using a data-driven approach.

Analyzing the flow fields around airfoils using the so-called Navier-Stokes PDE to represent the physical system is a classical use case in CFD. The traditional CFD solvers compute the velocity field and the pressure of the flow around the airfoil, enabling the understanding and improvement of airfoil properties and designs. 
The simplified Reynolds-averaged Navier-Stokes (RANS) equations \cite{reynolds1895iv} are time-averaged equations of motion for fluid flows that are commonly used for airfoil simulations. The system modeled by RANS equations can be expressed as:
\begin{align} \label{eq:RANs}
    \frac{\partial{u_i}}{\partial x_i} = 0 \\
    u_j \frac{\partial{u_i}}{\partial x_j} = \frac{\partial}{\partial x_j} [-p \delta_{ij} + (v+v_t)(\frac{\partial{u_i}}{\partial x_j} + \frac{\partial{u_j}}{\partial x_i})]
\end{align}
where $U = (u_x, u_y)$ is the velocity field and $p$ the pressure, the unknown quantities to be determined. $\nu$ is the fluid viscosity considered as a constant, and $\nu_t$ is the so-called eddy viscosity.
$\nu_t$ can be determined using the widely-adopted Spalart-Allmaras equation \cite{spalart1992one}.  
 RANS equations of four partial differential equations in a two-dimensional space are obtained by integrating the Spalart-Allmarasa equation. 
 A no-slip boundary condition is imposed along the wall of the airfoil, meaning that U is set to zero along all the points at the wall.
 To solve this system, the Semi-Implicit Method for Pressure-Linked Equations (SIMPLE) algorithm is commonly employed \cite{ferziger2002computational}, operating under the assumption of incompressible fluid behavior. 

\subsubsection{Machine Learning Scenario}
Real-world applications of airfoil flow simulation involve studying various airfoil shapes to optimize aircraft design for maximum energy capture and efficiency. 
In such context, our objective is to construct a model able to predict in short time  fluid flow (velocity and pressure) around an airfoil based on its shape (and an associated mesh) and the two flow parameters, Angle of Attack (AoA) and Mach number (Mach), describing the specificities of the physical system.
To achieve this, we will train a GNN model using a dataset generated by a numerical CFD solver, encompassing a variety of triplets (airfoil shapes, AoA, Mach).

It is critical that the trained model provides accurate predictions for airfoil shapes not seen during the training phase or those that significantly differ from the training set (out-of-distribution -- OoD) in real-world applications in aeronautics. 
However, traditional supervised learning algorithms discussed in Section \ref{sec: data-driven} encounter difficulties in achieving this objective (see e.g., the results of the baseline in Table \ref{tab:resultsMAML}). Thus, we propose utilizing meta-learning techniques as a solution to address this issue, enabling the trained model to maintain the high performance for OoD problems.   

\section{Meta-Learning} 
Meta-learning \cite{hospedales2021meta}, aka  "learning to learn", has emerged as a promising approach for training models that efficiently adapt to new tasks outside the ones used during training. In the context of multi-task learning, whereas standard machine learning algorithms are concerned with solving one single task, meta-learning involves learning, over several different tasks, some incomplete model that is able to learn a specialized model for all the tasks in training set from only a few data points for each task. This incomplete model is then able to also learn new tasks with only a few data points. 

\subsection{Problem Set-Up} 

A Meta-Learning dataset $\mathcal{D}_{meta}$ is a set of different supervised learning tasks: $\mathcal{D}_{meta} = \{\mathcal{T}_i;  i=1,\ldots,N \}$. Each task ${\cal T}_i$ is defined by a dataset of labeled sample points $\mathcal D_i = \{(x_1^i, y_1^i), \ldots, (x_k^i, y_k^i) \}$ containing a limited number of examples for the problem at hand, 
and a loss function $\mathcal{L}_i$ to be minimized: $\mathcal{T}_i = (\mathcal D_i, \mathcal L_i)$. 
As usual in supervised learning, each dataset $\mathcal{D}_i$ is split into a training and a test set: $\mathcal{D}_i = (\mathcal{D}_i^{tr},  \mathcal{D}_i^{test})$. Similarly, the meta-dataset is partitioned into a meta-training and a meta-test set: $\mathcal{D}_{meta} = (\mathcal{D}_{meta}^{tr},  \mathcal{D}_{meta}^{test})$. 

The goal of meta-learning is to train a model (aka meta-learner)  that can quickly adapt to all the tasks in  $\mathcal{D}_{meta}^{tr}$. This meta-learner is then tested on the tasks in $\mathcal{D}_{meta}^{test}$. The underlying reasoning is that if it generalizes well on the tasks in $\mathcal{D}_{meta}^{tr}$ using only a small number of specific examples, it should later similarly perform well on any new task, using, again, only a small number of examples.

\subsection{Model-Agnostic Meta-Learning} 
\label{MAML}

The model-agnostic meta-learning (MAML) \cite{finn2017maml} is a widely used method in meta-learning. 
The aim of MAML is to learn some initial parameters $\theta_0$ of a parameterized NN model $f_{\theta}$, such that $f_{\theta_0}$ can quickly be fine-tuned to any task in the meta-training set $\mathcal{D}_{meta}^{tr}$,  using a small number of examples of that task. 
Given $f$ parameterized by $\theta$, we will denote $\mathcal L(\mathcal D, \theta) = \sum_{(x,y)\in \mathcal D} \mathcal L(f_{\theta}(x), y)$

\subsubsection{Meta-Training} 
All details can be found in \cite{finn2017maml}, but a schematic view with on gradient step for meta-tuning is pseudo-coded in Algorithm \ref{alg:MAML}, in which $\mathcal D_{meta}^{tr}$ is identified to the set of airfoil shapes $\{\mathcal A_1, \ldots, \mathcal A_n\}$. 

Two optimization processes are embedded, the upper one updating the parameter $\theta_0$ of the meta-learner, and the inner one performing a few gradient descent steps on each task to evaluate the ability to quickly learn on the different tasks of $\mathcal D_{meta}^{tr}$.

\paragraph{Inner loop}: From the current meta-learner $f_{\theta_0}$, $M$ gradient descent steps are performed for each task $\mathcal T_i \in \mathcal D_{meta}^{tr}$, leading to updated values $\theta_i^k$ at the different steps $k \in [1,M]$, using the examples in $\mathcal D_i^{tr}$ (Line 9 of Algorithm 1, with $M=1$ and $k$ omitted). 

\paragraph{Outer loop}: The update of $\theta_0$ is then done by one gradient step too, aiming at minimizing the test error on all tasks $\mathcal T_i \in \mathcal D_{meta}^{tr}$, i.e., the error on the corresponding test set $\mathcal D_i^{test}$ (Line 11 of Algorithm 1). 


\subsubsection{Meta-Test} During Meta Test, given a new meta-test task $\mathcal{T} = ((\mathcal{D}^{tr}, \mathcal{D}^{test}), \mathcal{L})$ sampled from the same meta-distribution as the training tasks $\mathcal{T}_1,\ldots,{\cal T}_N$, the task-specific parameters $\theta$ for the new task $\cal T$ is updated using the same fine-tuning method as in the inner loop of the training process. The performance of the whole meta-learning algorithm is then evaluated on the meta-test set $\mathcal{D}_{meta}^{test}$ (and possibly for several different new meta-test tasks $\cal T$).

\subsubsection{MAML++} In fact, the update of the meta-parameters $\theta_0$ requires second-order derivatives. As a result, the bi-level optimization problem introduces instabilities when training the meta-learner. In order to mitigate these instabilities, \cite{maml++} proposed a multi-step loss optimization method (MSL). Instead of minimizing only the loss from the last step, the idea is to minimize the weighted sum of the losses at every updating step j, where the weights $w_j$ are pre-defined by users (see \cite{maml++} for all details and related experiments).
 Compared to the original MAML method, the modified version MAML++ stabilizes the training process, facilitating quicker and easier convergence. From now on, by abuse of notation, we will simply write MAML for MAML or MAML++ indistinctively. However, all experiments in Section \ref{sec:exp} have been performed using MAML++.
 
\subsubsection{MAML for PDEs} Nowadays, there has been a growing trend in applying meta-learning algorithms to the approximation of PDEs \cite{qin2022meta,huang2022meta,chen2022meta}. These approaches often integrate MAML and its variant \cite{reptile} with unsupervised learning methods for PDEs, such as physics-informed neural networks \cite{raissi2019physics} (PINNs). Researchers consider related parametric PDEs  as a set of related but distinct tasks and employ meta-learning principles to learn an initial model with strong generalization capabilities for these PDEs, thereby reducing the number of iterations required to adapt to a new PDE task. Such works partially reduce the computational cost for PINNs when a large number of PDEs with different parameters are to be solved but are nevertheless still much less efficient than data-based approaches. 

\subsection{Meta-Learning for Airfoil Flow}
Experiments on data-based approaches to airflow simulations have demonstrated that, whereas the generalization with respect to AoA and Mach is acceptable, that with respect to the airfoil shape is not, even more so when the shapes are chosen out of the distributions of the shapes in the training set: the results in Table \ref{tab:resultsMAML} are the visible part of these results, not shown here due to space constraints. Hence we propose to consider different airfoil shapes as different tasks, and to use MAML to train a meta-learner capable of adapting to new shapes with only a few sample data points for each shape. 
\subsubsection{Rationale} Transfer learning is useful for enhancing the predictive accuracy of the trained model, in particular regarding OoD shapes. Given an OoD airfoil, transfer learning can be simply employed by fine-tuning a pre-trained model on a small dataset of pairwise data from that OoD airfoil.  
This approach has a low computational cost on the training process and helps alleviate overfitting concerns. However, transfer learning may be less effective when limited data (less than 100 examples) is available for the new task \cite{UniversalLanguage}. As a result, collecting data for fine-tuning a model on OoD airfoils can be resource-intensive, requiring running numerical simulations hundreds of times to generate new problem instances for each airfoil. This can be burdensome, especially when the model needs to be fine-tuned for multiple OoD airfoils. 

\begin{figure}[ht]
    \centering
    \includegraphics[scale=0.45]{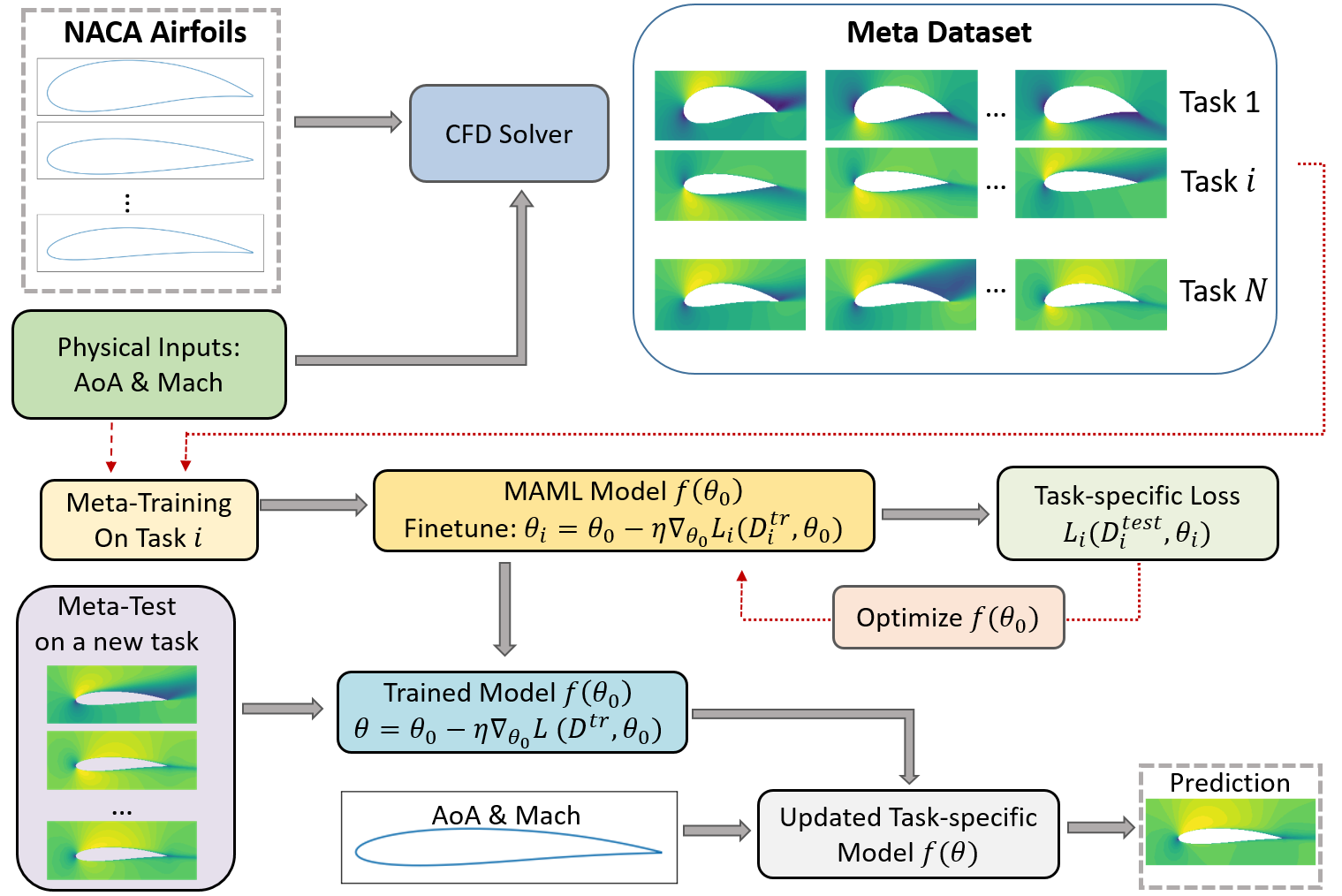}
    \caption{A schematic view of MAML algorithm to solve Airfoil Flow Problems}
    \label{fig:diagram}
\end{figure}

Instead of transfer learning from a pre-trained model for each new airfoil shape, it is natural to consider employing the MAML algorithm, which embeds the fine-tuning process within the training phase, to develop a meta-learner proficient in fine-tuning for novel airfoil shapes. This approach can reduce the requirement for additional data points.
\\
\\
To accomplish this, we formulate the airflow problem over various airfoils as a meta-learning problem (see Figure \ref{fig:diagram}), where each set of examples defined on a single airfoil shape is treated as a separate task. The modified version MAML++ is used to learn an incomplete model $f_{\theta_0}$, to be fine-tuned with a few steps of gradient descent, for any task, both from the meta-training set and the meta-test set. 
\begin{algorithm}[ht]
    \caption{Model-Agnostic Meta-learning Algorithm for Airfoil Flow}
    \label{alg:MAML}
        \begin{algorithmic}[1] 
        \Require A set of airfoil shapes $\{\mathcal{A}_1, ..., \mathcal{A}_n \}$ from the shape distribution.
        \State Set learning rate parameters $\eta_1$, $\eta_2$
        \State Initialize the meta parameters $\theta_0$

        \For{every training airfoil $\mathcal{A}_i$ }
            \State Sample disjoint dataset $(D_i^{tr}, D_i^{test})$ from $\mathcal{A}_i$ by sampling AoA and Mach. 
        \EndFor

        \While {Not Converged}
        \For{every training airfoil $\mathcal{A}_i$ }
            \State Evaluate gradient descent on $\theta$ with $D_i^{tr}$
            \State Compute adapted parameters
            $\theta_i = \theta_0 - \eta_1 \nabla_{\theta_0} \mathcal{L}_i (D_i^{tr}, \theta_0)$
        \EndFor
        \State Update the meta-parameters $\theta_0 = \theta_0 - \eta_2  \sum_{\mathcal{T}_i \in \mathcal D_{meta}^{tr}} \nabla_{\theta_0}\mathcal{L}_i( D_i^{test}, \theta_i)$ 
        \EndWhile
        \end{algorithmic}
\end{algorithm} 

\section{Experiment}
\label{sec:exp}
\subsection{Data Preparation} 
\label{dataPreparation}
We generate a database of cases with various angles of attack (AoA), Mach numbers, and airfoil shapes as described below. 
The CFD solver $\texttt{OpenFOAM}$ \cite{weller1998tensorial} is applied to generate 
ground truth on structured C-grid meshes in our work.

\subsubsection{Airfoil Generation}
We create 80 distinct NACA 4-digit airfoil shapes characterized by their camber $C$, the position of their maximum camber $P$, and their thickness $T$. All three parameters are randomly sampled from uniform distributions: $C \sim \mathcal{U}[0, 0.09]$, $P \sim \mathcal{U}[0.4, 0.6]$ and $T \sim \mathcal{U} [0.1, 0.3]$. 

\subsubsection{Example Generation}
We construct C-grid meshes for each NACA airfoil using an automated algorithm \cite{curiosityMesh} for further ground truth generation with the CFD solver $\texttt{OpenFOAM}$.
For each NACA airfoil, 40 examples are generated by sampling uniformly the two physical parameters, Angle-Of-Attack(AoA) in [-22.5, 22.5] (in degrees), and Mach number in [0.03, 0.3].

\subsubsection{Data Pre-processing}
We follow the recommended data pre-processing steps outlined in \cite{thuerey2020deep}, concentrating on a small region near the airfoil and normalizing the output quantities.
Airflow simulations by CFD solvers require a computational domain that is much larger than the airfoil so as to maintain an adequate distance between the airfoil object and the boundaries of the computational domain, to ensure that the boundary conditions assigned to the outer domain do not impact the quality of the flow simulation around the airfoil. 
However, when training the NN models, it is not necessary to compute predictions for areas far away from the airfoil, which is of less interest (and remains almost constant, equal to the boundary conditions). As a result, we focus the learning on a small domain in close proximity to the airfoil.
Additionally, we normalize the output quantities, velocity $U$ and pressure $p$, see Section \ref{CFDdimulations}, relative to the magnitude of the freestream velocity $U_0$ to make these quantities dimensionless: 
$$\bar{U} = {U}/{||U_0||}, \quad \bar{p} = {p}/{||U_0||^2}$$
\subsubsection{Network Inputs and Outputs} 
Both input and output are represented in the form of graph data. Formally, graph data can be expressed as $G = (N, E, {\cal V}, {\cal E})$,  where $N$ is the graph nodes, $E$ the set of edges, ${\cal V} \in \mathbb{R}^{N \times F}$ are the $F$-dimensional node attributes attached to each node, and ${\cal E} \in \mathbb{R}^{E \times D}$ the D-dimensional attributes attached to each edge. 

In our case, the CFD solver $\texttt{OpenFOAM}$ uses finite volume methods to discretize the PDE. The simulation results provide an approximate value of the output quantities at the center of each cell. 
Rather than preserving the mesh nodes, the centroids of each cell are converted to the nodes of the graph, and two centroids $i$ and $j$ are connected if their corresponding cells are adjacent (i.e., share an edge). Moreover, the edge attribute $x_{ij}^e \in {\cal E} $ of edge ($i,j$) is $ x_{ij}^e = c_j - c_i $, where $c_i$ and $c_j$ are the coordinates of nodes $i$ and $j$ respectively (D=2), representing the distance relations in the mesh.
The initial freestream condition $U_0 = (v_{0, x}, v_{0, y})$ depending on AoA and Mach is considered as the input node attributes (F=2). 
The output node attributes are the velocity and pressure at node i (F=3).   

\subsection{Model Architecture} \label{sec: graph_unet}
The model architecture employs the Graph U-Net architecture proposed in \cite{liu2021multi}, which combines the advantages of Graph Neural Networks (GNNs) and the U-Net architecture \cite{unet}. This Section will briefly give its main characteristics for the sake of completeness.
\subsubsection{Graph Neural Networks}
Graph Neural Networks (GNNs) \cite{bruna2014spectral,defferrard2017convolutional,kipf2017semisupervised,atwood2016diffusionconvolutional,velickovic2018graph} are designed to treat data from non-Euclidean spaces, such as graphs, meshes, or manifolds, and are now widely used for processing mesh data in the physical domain \cite{kashefi2021point,belbuteperes2020combining,pfaff2020learning}. Key components of a GNN include a convolutional operator that is invariant to node permutations.
\\
Among the different types of GNN, we use the MoNet GNN \cite{monti2016geometric}, defined as follows. Consider a weighted graph  $G=(N, E, {\cal V}, {\cal E})$.
Let $x_i \in \mathbb{R}^F$ be the feature vector of node $i$, and $x^e_{ij} \in \mathbb{R}^D$ be the feature of the edge $i, j$ defining the set of neighbors $N(i)$ of node $i$. The basic idea of MoNet is to define a trainable function $\mathbf{w}$ that computes an edge weight $w_{ij}$ from the edge feature $x^e_{ij}$.
MoNet then defines the convolutional operator at node $i$ as:
\begin{equation*}
\mathbf{x}^{\prime}_i = \frac{1}{|\mathcal{N}(i)|}
        \sum_{j \in \mathcal{N}(i)} \frac{1}{K} \sum_{k=1}^K
        \mathbf{w}_k(\mathbf{x}^e_{ij}) \odot \mathbf{\Theta}_k \mathbf{x}_j
\end{equation*}
where $\mathbf{K}$ is the user-defined kernel size,
$\mathbf{\Theta_k} \in \mathbb{R}^{M \times N}$ stands for the trainable matrix applying a linear transformation on the input data,
$\odot$ is the element-wise product,
and $\mathbf{w}_{k}, k=1,\ldots,K$ are trainable edge weights. Following \cite{monti2016geometric}, we use Gaussian kernels:
$\mathbf{w}_{k}(x^e_{ij}) = exp(-\frac{1}{2}(x^e_{ij}-\mu_k)^{T} \Sigma_k^{-1}(x^e_{ij} - \mu_k))$.
Both $\mu_k$ and $\Sigma_k$ are trainable variables.

\subsubsection{Graph U-Net}
GNNs can directly handle mesh data, making them well-suited for processing complex geometries like airfoil shapes. Meanwhile, the U-Net architecture \cite{unet} is known in image processing for its ability to extract hierarchical spatial features, which helps in capturing essential information at various scales.
Modern mesh generators enable the creation of hierarchical meshes for the same domain while preserving geometric information. The model leverages these algorithms to build a hierarchy of meshes and construct the Graph U-Net architecture. A series of graph layers are applied at the same mesh level, while pooling operators transfer data between different mesh levels.

\subsubsection{Pooling Operator}
The k-nearest interpolation proposed in PointNet++ \cite{qi2017pointnet2} was chosen to design the pooling operators. It is consistent with standard interpolation routinely used in piecewise continuous numerical approximations. Let $y$ be a node from ${\cal M}_1$, and assume its $k$ nearest neighbors on ${\cal M}_2$ are $(x_1, \ldots, x_k)$. The interpolated feature for $y$ is defined from those of the $x_i$'s as:
$$\mathbf{f(y)} = \frac{\sum_{i=1}^k w(x_i) \mathbf{f}(x_i)}{\sum_{i=1}^k
        w(x_i)} \textrm{, where } w(x_i) = \frac{1}{||y - x_i||_2}$$

\subsection{Training the Meta-Learner} \label{sec:Meta-Train}

The Meta-Learner is trained based on our described methodology in Section \ref{MAML}. A 10-fold cross-validation is used to assess the robustness of the proposed approach. 
The loss function is defined as the Mean Squared Error (MSE) between the prediction and ground truth. Our approach is compared with two baselines: The first baseline uses a global (single-task) approach, and considers the multi-datasets as one single  dataset, to which it applies standard supervised learning. 
MAML models require additional data every time a new task is presented. In contrast, the baseline model is applied directly to unseen airfoils. 
However, such extra data (the few labeled examples available for the new tasks/airfoils) can help not only MAML but maybe also the baseline model: To be fair to the baseline,  the second baseline consist in fine-tuning it at test time, for each new task, using the same additional data as MAML, called "Baseline+FT". 
10-fold cross-validation is used in all experiments and means and standard deviations on the (meta-) test sets are reported.

\subsubsection{Hyper-parameters Setting} In order to validate the effectiveness of the proposed approach and ensure a fair comparison, the architecture and hyperparameters of the Meta-Learner are the same as those that were selected for the baseline model. The graph U-Net architecture detailed in Section \ref{sec: graph_unet} is used, with the same graph blocks and sampling operators. Each graph block $C$ contains multiple graph layers followed by the activation function "ELU." The block $C$ is specified by the number of layers $l$, a channel factor $c$, and a kernel size $k$. A sampling operator is applied after each GNN block to transform data between two mesh levels. The sampling operator has one hyper-parameter, the number of nearest neighbors, set to 6. The model down-samples progressively three times in the encoding part and recovers the initial mesh resolution with three up-sampling operators during decoding, containing seven GNN blocks $C = (2, 48, 5)$.
Finally, a graph layer maps the high-dimensional features to the solution space. 
During meta-training, the Adam optimizer is used with an initial learning rate of 5E-4 and step decay by a factor of 0.5 every 500 epochs.
The number of inner gradient update steps is set to 3, and the first 20 examples of each task $\mathcal{T}_i$ are considered as the train set $\mathcal{D}_i^{tr}$ used for inner gradient updates with a fixed learning rate of 0.01, while the remaining examples are used as $\mathcal{D}_i^{test}$ to compute the meta-loss $\cal L$ for meta-update.

\subsection{Experimental Results} \label{results}
\subsubsection{Meta-Test Sets}
Three meta-test sets allow us to evaluate the generalization properties of the models with respect to airfoil shapes. For the flow parameters AoA and Mach number, we use the same distribution as in the training set in order to more accurately highlight the performance of the models on OoD tasks (see Section \ref{dataPreparation}).
\begin{itemize}
    \item \textbf{Flow Interpolation Set}: As a sanity check, we generate 10 additional examples (AoA, Mach) for each airfoil shape (meta-training task). For evaluation, we directly use the initial training data to update each task and make predictions for new examples. 
    \item \textbf{Shape Interpolation Set}: 
    This set includes 20 new airfoil shapes that are generated using the same distributions of NACA parameters as the training set (see Section \ref{dataPreparation}).  Each airfoil is treated as a separate meta-test task, with 50 (AoA, Mach) examples in total. We use the first 20 such examples to update the Meta Learner, and the remaining 30 examples to evaluate the model performance.
    \item \textbf{Out of Distribution Set}: Thinner and less regular airfoil shapes are created by altering the distribution range for NACA parameters $T$ and $P$ ($P \in [20, 80]$ and $T \in [5, 10]$). Similarly to the Shape Interpolation Set, 50 (AoA, Mach) examples are created for this meta-test set, that are more challenging than those in the training set.
\end{itemize}
Unless otherwise stated, our meta-model is evaluated after fine-tuning on each meta-test task with 10 gradient updates and compared with the baseline model on the three meta-test sets described above.
\subsubsection{First Results}
We use  RMSE as the metric for model performance and compare it with the two baselines outlined in Section \ref{sec:Meta-Train}. Moreover, for all pair-wise comparisons between test errors on MAML and Baseline, we performed a Wilcoxon signed-rank statistical test with 95\% confidence, and the differences between all pairs are statistically significant.
\begin{table}[h!] 
  \caption{Evaluation Results on different test sets with 10 gradient updates}
  \centering
\begin{tabular}{lccc}  
\toprule
&\multicolumn{3}{c}{RMSE(1e-2)} \\
\cmidrule(l){2-4}
Test Sets         & MAML                & Baseline              & Baseline + FT         \\
\midrule
Flow Interpolation   & 0.79 $\pm$ 0.08 & \textbf{0.71 $\pm$ 0.03} & 0.71 $\pm$ 0.03 \\
Shape Interpolation   & \textbf{1.18 $\pm$ 0.14} & 1.58 $\pm$ 0.15 & 1.48 $\pm$ 0.14 \\ 
Out-of-Distribution & \textbf{3.69 $\pm$ 0.28} & 6.43 $\pm$ 1.79 & 5.35$\pm$1.62        \\        
\bottomrule
\end{tabular}
\label{tab:resultsMAML}
\end{table} 

From the results in Table \ref{tab:resultsMAML}, it turns out the MAML model outperforms the baseline on the Shape Interpolation and OoD meta-test sets, while, as expected, the baseline performs best on the Flow Interpolation set -- and fine-tuning of course does not improve its performance. By using just a few data points and gradient steps, MAML models can quickly adapt to new tasks without overfitting. 
Interestingly, adding some fine-tuning to the baseline does improve its performance on the Shape Interpolation and OoD meta-test sets, but the results remain far below those of the MAML approach.
\begin{figure}[h]
    \centering
    \includegraphics[scale=0.5]{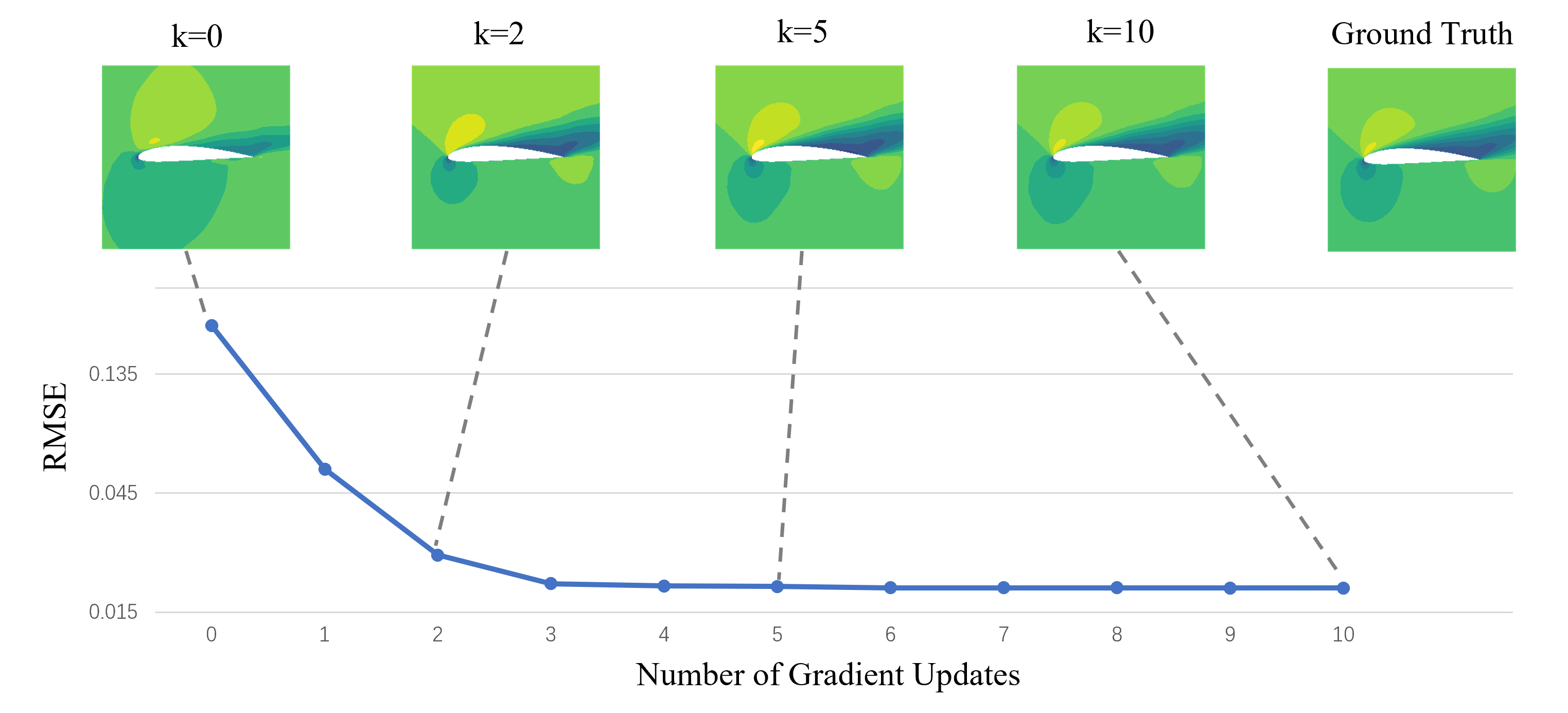}
    \caption{An example of prediction from MAML approach. By observing the intermediate plots and loss, the Meta-Learner demonstrates rapid convergence toward the ground truth }
    \label{fig:my_label}
\end{figure}

\subsubsection{Number of gradient updates}  
\begin{figure}[h]
    \centering
    \includegraphics[scale=0.4]{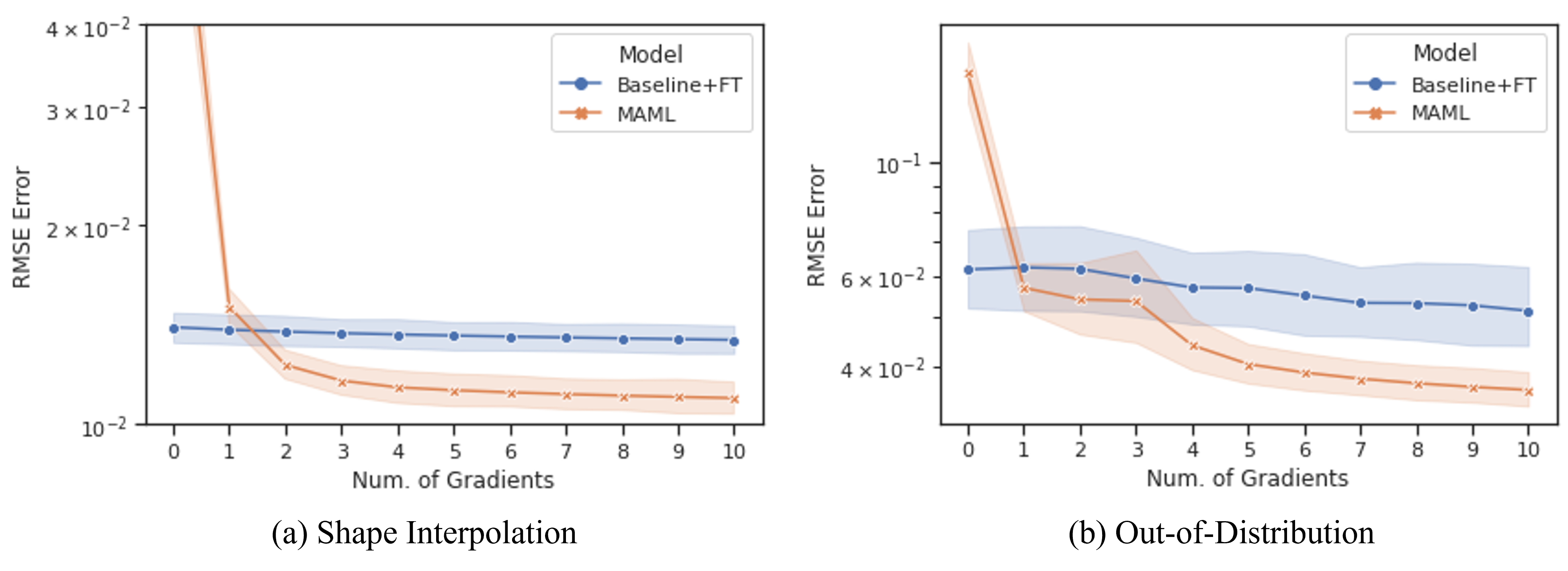}
    \caption{Sensitivity w.r.t. the number of gradients . The MAML model continuously improves, while the baseline does not show such significant improvements.}
    \label{fig:gradients}
\end{figure}

Figure \ref{fig:gradients} displays plots of the RMSE error w.r.t. the number of gradient updates during the Meta-Test on the Shape Interpolation and OoD meta-test sets. It shows that, given very limited data, the model trained with MAML can rapidly adapt to novel tasks and continue to enhance performance without overfitting. 
Moreover, the figure highlights a comparison between the baseline with fine-tuning approach and the meta-trained model. While fine-tuning does improve the performance of the baseline model, the meta-trained model shows more significant improvements in performance. Finally, we observe on the OoD meta-test set that the decrease of the error is steeper and lasts longer for the meta-trained model, with a much lower variance than for the baseline. 

\subsubsection{Number of examples in meta-test sets} 
As previously mentioned, during the meta-training phase, 20 examples were employed for inner gradient updates. At the meta-test time, the same number of examples is used to update the meta-learner for meta-test tasks in the first experiment.
\\
We now investigate the impact of the number of sample data points used for meta-test tasks on the performance of the MAML model: The MAML model is fine-tuned using from 1 to 20 examples, each with 10 gradient updates, in order to understand how the number of examples used for test tasks affects the model performance.
\begin{figure}[h]
    \centering
    \includegraphics[scale=0.4]{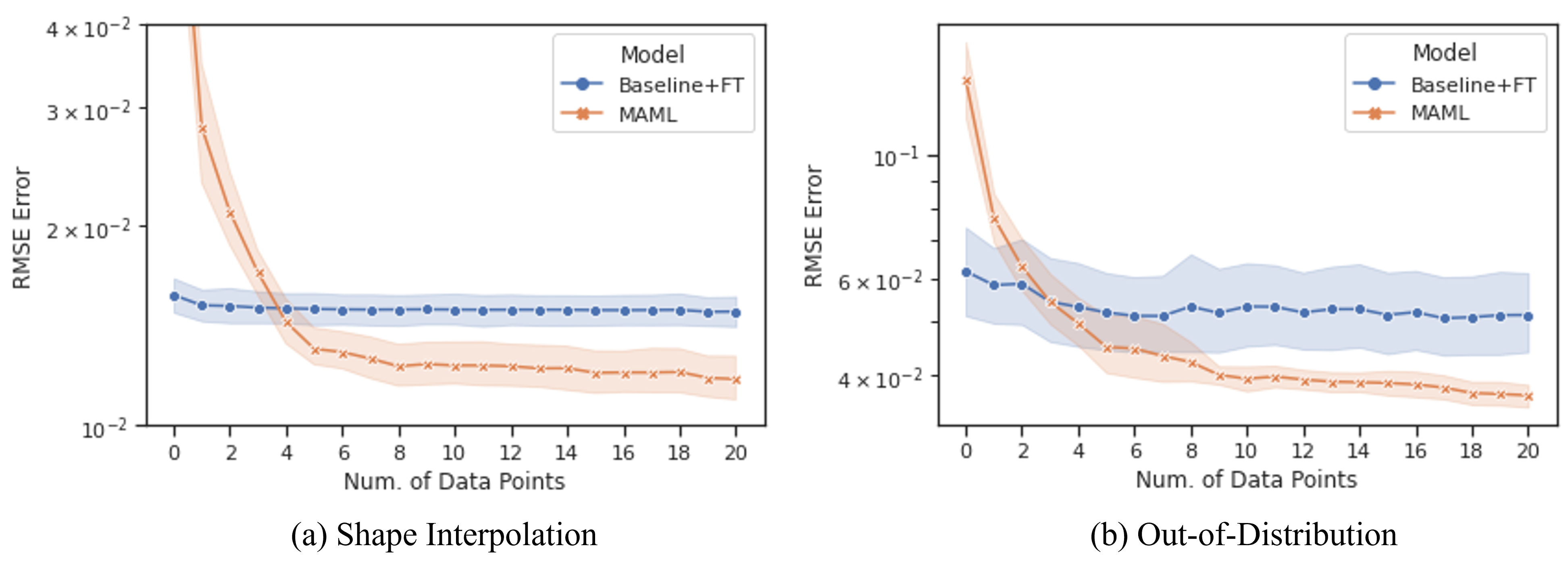}
    \caption{RMSE w.r.t. number of examples}
    \label{fig:examples}
\end{figure}

From Figure \ref{fig:examples}, increasing the number of examples used to update the meta-learner has a positive impact on the model performance. While the MAML model still performs well on new tasks and outperforms the baseline model when using more than 5 examples, it is unable to make reasonable predictions (i.e., at least as good as the baseline) on new meta-tasks with using less than 4 examples: At least 4 examples are necessary for the Shape Interpolation set, while 3 are sufficient on the OoD set, where the baseline is far less efficient. 

\section{Conclusion}
In this paper, we have presented a meta-learning approach to address airfoil flow problems. By utilizing the Model-Agnostic Meta-Learning (MAML) algorithm, we have trained a meta-learner that is able to adapt to new tasks with only a small number of examples.

Our experimental results show that the model obtained by meta-learning consistently outperforms the baseline model (a straightforward single-task approach which is trained classically, once and for all using the whole dataset) in terms of its ability to generalize to different airfoil shapes, even more so for out-of-distribution shapes -- and this, even if this baseline model is fine-tuned with the same small amount of data for the new tasks than the MAML meta-model. 

One drawback of the meta-learning approach is the need for a few examples for any new task. But the clear benefits in performance should justify it in most practical cases.

Overall, we claim that the meta-learning approach represents a promising solution for addressing the challenges of generalization in airfoil flow problems. Furthermore, this approach is not specific to airflow problems and could be used on top of any complex and computationally costly numerical simulation.

\bibliographystyle{splncs04}
\bibliography{reference}

\end{document}